\pdfoutput=1

\documentclass[11pt]{article}

\usepackage[final]{acl}

\usepackage{times}
\usepackage{latexsym}

\usepackage[T1]{fontenc}
\usepackage{pifont}

\usepackage[utf8]{inputenc}

\usepackage{microtype}

\usepackage{inconsolata}

\usepackage{csquotes}
\usepackage{graphicx}
\usepackage{enumitem}
\usepackage{mdframed}
\usepackage{amsmath}
\usepackage{indentfirst}
\usepackage{multirow}
\usepackage{algorithm}
\usepackage{algorithmic}
\usepackage{tabularx}

\usepackage{amssymb}

\title{\textit{Take the Essence and Discard the Dross}: \\
A Rethinking on Data Selection for Fine-Tuning Large Language Models}

\author{
Ziche Liu\textsuperscript{1}\thanks{The first two authors are equal contributions.}, 
Rui Ke\textsuperscript{1}$^*$, 
Yajiao Liu\textsuperscript{1},
Feng Jiang\textsuperscript{1,2,3}\thanks{Corresponding Author.}, 
Haizhou Li\textsuperscript{1,2}\\
  \textsuperscript{1}The School of Data Science, The Chinese University of Hong Kong, Shenzhen \\
  \textsuperscript{2}Shenzhen Research Institute of Big Data \\
  \textsuperscript{3}University of Science and Technology of China \\
  \texttt{jeffreyjiang@cuhk.edu.cn} 
  }

\begin{document}
\maketitle
\begin{abstract}
Data selection for fine-tuning large language models (LLMs) aims to choose a high-quality subset from existing datasets, allowing the trained model to outperform baselines trained on the full dataset. 
However, the expanding body of research lacks a clear, unified framework, and the variability in experimental settings complicates systematic comparisons.
While existing surveys comprehensively overview the stages and methods of data selection, they often overlook an in-depth exploration of the fine-tuning phase. In this paper, we conduct a focused review of recent data selection techniques for fine-tuning LLMs, analyzing a dozen key studies. We introduce a novel three-stage scheme—comprising feature extraction, criteria design, and selector evaluation—to systematically categorize and evaluate these methods. Additionally, we propose a unified comparison approach that incorporates ratio-based efficiency and ranking-based feasibility metrics to address inconsistencies across experiments. Our findings reveal that methods emphasizing more targeted quality measurement achieve higher efficiency but at the cost of feasibility. Finally, we discuss trends and highlight four key challenges in fine-tuning data selection, offering potential directions for future research~\footnote{The code is available at \url{https://github.com/tREeFrOGcoder/TEDD-Ranker}}.
\end{abstract}

\section{Introduction}
\label{sec:intro}

\begin{figure}[!t]
\centering
\includegraphics[width=\linewidth]{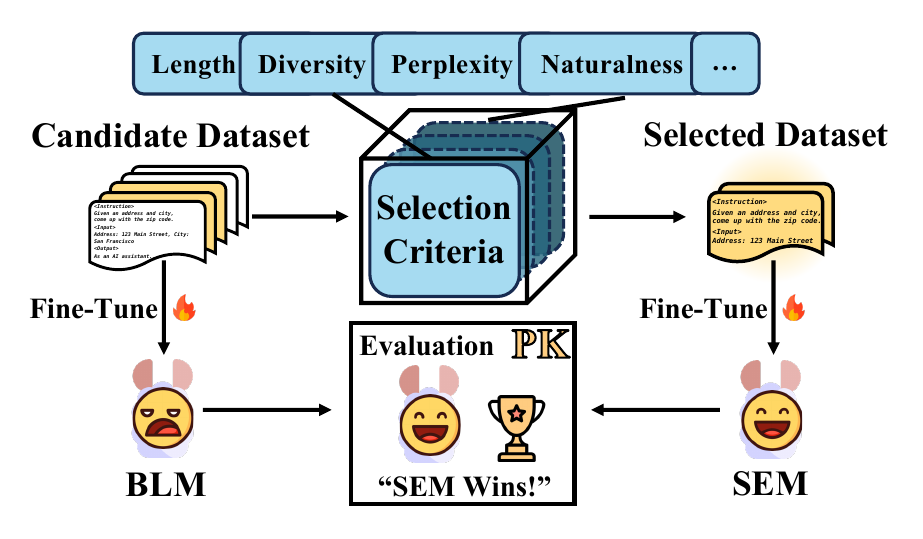}
\caption{An illustration of data selection for fine-tuning LLMs. 
Fine-tuning a model on the full dataset results in a BaseLine model (BLM), while training a model on a selected high-quality subset produces the Selective-Enhanced Model (SEM), which is expected to outperform the BLM.}
\label{fig:one-pic}
\end{figure}

Supervised fine-tuning (SFT) leverages small amounts of instruction-pair data to unlock large language models' instruction-following capabilities and improve generalization across various tasks \cite{radford2019language, weifinetuned, singh2023beyond, zhang2024instructiontuninglargelanguage, albalak2024surveydataselectionlanguage}. Recent research highlights that data quality is more critical than data quantity for effective fine-tuning~\cite{nakkiran2021deep, shumailov2024ai, zhou2024lima, jindal2024birbal}. As a result, several data curation techniques have been proposed, such as data selection~\cite{chen2024alpagasus, li2024quantity}, data evolution~\cite{wang2023selfinstruct, xu2023wizardlm}, and data reflection~\cite{mukherjee2023orca, yin-etal-2023-dynosaur}. Data selection, in particular, involves choosing a high-quality subset from a candidate dataset based on specific selection criteria, enhancing the model's performance while improving training efficiency by reducing the number of samples. Unlike data augmentation or polishing, it focuses on selecting inherently higher-quality samples, as shown in Figure \ref{fig:one-pic}.

\begin{figure*}[!t]
\centering
\includegraphics[width=\textwidth]{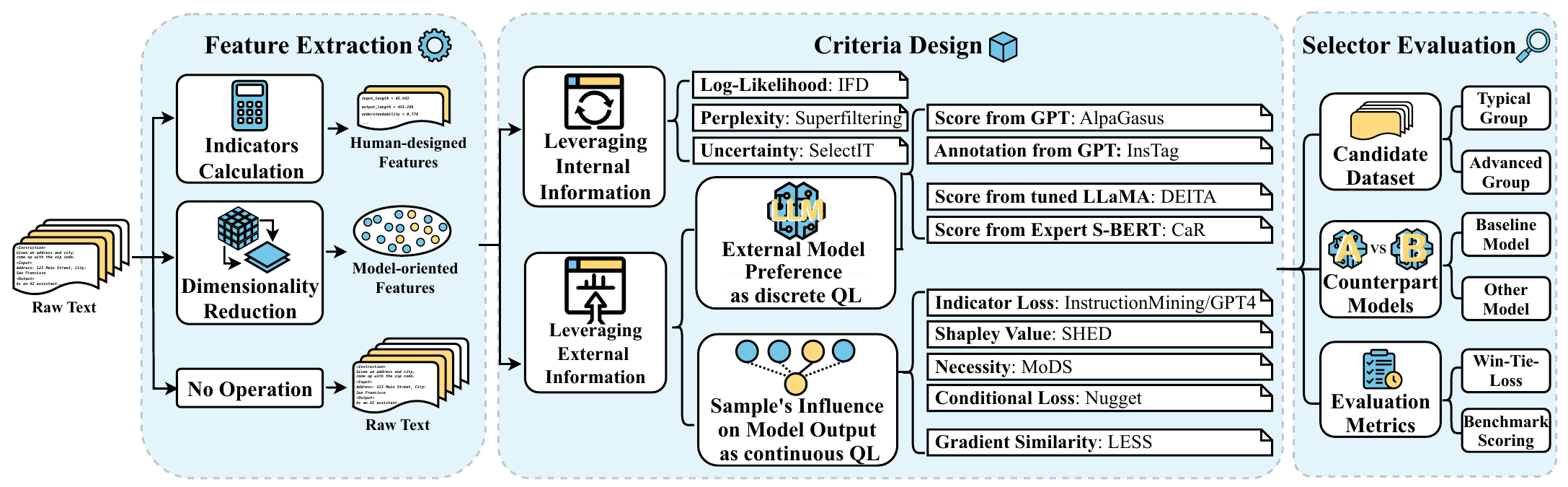}
\caption{The Three-stage Scheme of Data Selection for Fine-tuning LLMs. 
The feature extraction stage transforms the raw data into compact representations to facilitate selection. 
The criteria design stage constructs quality labels (QL) to capture data quality as selection criteria. The selector evaluation stage identifies the key components to evaluate the effectiveness of the selector.
}
\label{fig:landscape}
\end{figure*}

However, despite the rapid development of data selection methods, there is currently no unified framework for systematically guiding and comparing these methods, as experimental settings vary widely across studies. Although some surveys~\cite{bommasani2021opportunities, albalak2024surveydataselectionlanguage, wang2024survey} have reviewed data curation techniques in the contexts of pretraining, fine-tuning, and reinforcement learning, they generally provide high-level overviews and lack in-depth discussions of the fine-tuning stage. This gap makes it difficult for researchers to conduct focused, sustained studies on this crucial phase. To address this, our survey provides a fine-grained review of data selection methods for instruction fine-tuning LLMs, rethinking existing approaches, proposing a unified comparison method, and outlining key trends and challenges in the field.

We begin by reviewing existing data selection techniques, organizing them within a three-stage scheme based on the key components of the data selection pipeline: feature extraction, criteria design, and selector evaluation (Figure \ref{fig:landscape}). In the feature extraction stage (Section~\ref{sec:preprocess}), we categorize methods into three types based on the form of the candidate data: human-designed features, model-oriented features, and raw text. 
In the criteria design stage (Section~\ref{sec:selector}), we categorize methods based on the source of the sample quality label into two groups: internal information and external information. The latter is further divided into methods obtaining criteria from model preference or sample influence. In the selector evaluation stage (Section~\ref{sec:eval}), we outline three key aspects to reliably evaluate a selector's effectiveness: candidate datasets, counterpart models, and evaluation metrics.

We also introduce a unified comparison method for evaluating existing works, incorporating both ratio-based efficiency indicators and ranking-based feasibility indicators (Section \ref{sec:anal}). Specifically, we first construct a quantitative comparison of two-dimensional efficiency based on the performance improvement ratio (PIR) and selected dataset fraction (SDF), aligning them through the efficiency curve assumption, effectively addressing the challenge of comparing different methods under inconsistent configurations. Then, we consider the feasibility of the method from the perspectives of flexibility and simplicity indicators. It qualitatively ranks existing models by manually considering algorithm complexity and reproducibility (the number of training models, algorithm steps, and open-source availability), as well as their transferability and scalability (dependence on data and models). 

Finally, we discuss the main trends and challenges faced for data selection. We first sort out the existing works chronologically from three aspects (Candidate Dataset, Quality Measurement, and Selected Feature) to grasp the current research focus (Section \ref{sec:disc}). We then point out the most important open question (How can we design effective sample quality measurement for data selection?) with four challenges: balance the efficiency and feasibility; ensure the measurement objectivity; improve specific tasks/domains performance without compromising others and satisfy multiple goals.

\section{Feature Extraction}
\label{sec:preprocess}

Feature extraction plays a crucial role in the data selection process. Some works preserve the original text data, believing that it contains the most comprehensive information~\cite{chen2024alpagasus, li2024quantity}, while others transform raw texts into more compact and representative feature sets (e.g., human-designed features and model-oriented features) to facilitate selection. The former approach aligns with human instinct by retaining linguistic structures, while the latter focuses on extracting features directly from the model to provide more targeted data representations.

\textbf{Human-designed Features.} To guide data selection according to human preferences, certain methods employ explainable, human-designed features derived from linguistic information. For instance, Instruction-Mining~\cite{cao2023instruction} leverages such features by converting samples into vector representations that incorporate various NLP metrics, including coherence, naturalness, and understandability. These metrics reflect linguistic quality and can serve as effective selection criteria for improving the fine-tuning process.

\textbf{Model-oriented Features.} For more direct and targeted data selection, other works focus on model-oriented features, which are extracted directly from the trained model as data representations. These features aim to capture the essence of the model’s performance on specific tasks. For example, LESS~\cite{xia2024less} creates a datastore of effective and reusable low-dimensional gradient features from the LLM. This method allows the model to directly minimize the loss on a target task by selecting samples rather than relying on superficial or pre-defined features.

\section{Criteria Design}
\label{sec:selector}

The source of the quality label is a critical factor in designing selection criteria, which can be categorized into internal and external information. Internal information refers to quality signals inherent in the candidate dataset, while external information evaluates quality through external measures, such as preferences from external LLMs or the impact of samples on model performance.

\subsection{Leveraging Internal Information} \label{sec:internal}

To extract useful quality indicators from the given candidate dataset, early work~\cite{li2024quantity} introduced Instruction Following Difficulty (IFD) as a quality label, measuring how much an instruction contributes to generating the corresponding output. Specifically, this study first trains an LLaMA-2-7B model as a pre-experienced model on a subset of the target dataset. The IFD score of an instruction is then calculated based on how its presence affects the likelihood of the model generating a particular response compared to when no instruction is provided. Inspired by this, Superfiltering~\cite{li2024superfiltering} adopts a smaller model (e.g., GPT-2) as the pre-experienced model for data selection and leverages the consistency between IFD scores and perplexity across different model sizes to facilitate fine-tuning from weaker to stronger models. Additionally, SelectIT~\cite{liu2024selectit} employs a three-level reflection strategy (token-level, sentence-level, and model-level) to measure a backbone model's uncertainty about a candidate sample. This approach aims to reduce bias and improve the reliability of quality assessment.

\subsection{Leveraging External Information}
\label{sec:external}

Existing methods leveraging external information can be categorized into two approaches: external model preferences and sample influence on model outputs.

\subsubsection{External Model Preference as Discrete Quality Labels}
\label{sec:LLMPreference}

To address the inefficiency of relying on human labor for sample quality annotations, some works use exclusive LLMs (e.g., ChatGPT), community LLMs (e.g., LLaMA), or even smaller expert models to annotate quality automatically. These quality labels are typically derived from external model outputs and are, therefore, discrete.

\textbf{Exclusive LLM Preference.} AlpaGasus~\cite{chen2024alpagasus} is a notable example that prompts ChatGPT to score each sample directly, mimicking human evaluation. The prompt follows a structured template incorporating designed evaluation criteria, such as helpfulness and accuracy, making it adaptable across different datasets and backbone models. Unlike AlpaGasus, which assigns a single score per sample, InsTag~\cite{lu2023instag} generates fine-grained quality labels in the form of instruction intention tags annotated by ChatGPT. It evaluates sample quality based on diversity and complexity, and then applies a complexity-first, diversity-aware sampling algorithm to balance these factors in data selection.

\begin{table*}
    \centering
    \small
    \begin{tabularx}{\textwidth}{>{\raggedright\arraybackslash\hsize=0.6\hsize}X
                                  >{\raggedright\arraybackslash\hsize=1.3\hsize}X
                                  >{\raggedright\arraybackslash\hsize=1.1\hsize}X}
        \hline
        \textbf{Method} & \textbf{Candidate Datasets} & \textbf{Backbones of SEMs} \\
        \hline
        AlpaGasus & Alpaca, DOLLY, GPT-4-LLM & LLaMA-1/2-7B/13B\\ 
        Instruction-Mining & OpenOrca, DOLLY & LLaMA-1-7B, LLaMA-2-7B/13B \\
        CaR & Alpaca & LLaMA-1-7B/13B/30B, LLaMA-2-7B, LLaMA-3-8B \\
        SelectIT & Alpaca-GPT4 & LLaMA-2-7B/13B, LLaMA-3-8B/13B, Mistral-7B \\ 
        InstructionGPT-4 & cc\_sbu\_align & MiniGPT-4 \\
        IFD & Alpaca, WizardLM & LLaMA-1-7B, LLaMA-2-7B/13B \\ 
        Superfiltering & Alpaca, Alpaca-GPT4 & LLaMA-2-7B/13B \\
        Nuggets & Alpaca, CodeAlpaca, WizardLM, FLANv2  & LLaMA-2-7B \\  
        LESS & FLAN V2, CoT, DOLLY, Oasst & LLaMA-2-7B/13B, Mistral-7B \\
        SHED & MMLU, WizardLM & LLaMA-1-7B/13B, Vicuna-7B, GPT-2 \\ 
        MoDS & Alpaca, HC3, Alpaca-Evol-Instruct, DOLLY, InstructWild, LIMA & LLaMA-2-7B \\ 
        InsTag & WizardLM, UltraChat, ShareGPT & LLaMA-1/-2-13B\\
        DEITA & Alpaca, DOLLY, Oasst, FLAN 2022, WizardLM, UltraChat, ShareGPT & LLaMA-1/-2-13B, Mistral-7B \\ 
        \hline
    \end{tabularx}
    \caption{The candidate datasets and backbones of SEMs used in each method.}
    \label{tab:DataSum1}
\end{table*}

\begin{table*}[!ht]
    \centering
    \small
    \begin{tabularx}{\textwidth}{
        >{\raggedright\arraybackslash\hsize=1.05\hsize}X
        >{\centering\arraybackslash\hsize=0.22\hsize}X
        >{\centering\arraybackslash\hsize=0.38\hsize}X
        >{\raggedright\arraybackslash\hsize=1.4\hsize}X
        >{\raggedright\arraybackslash\hsize=1.95\hsize}X
    }
        \hline
        \textbf{Method} & \textbf{BLM} & \textbf{Others} & \textbf{Win-tie-loss} & \textbf{Benchmark Scoring} \\
        \hline
        AlpaGasus & \ding{52} & \ding{52} & Vicuna, Koala, WizardLM, Self-Instruct & MMLU, DROP, Humaneval, BBH \\ 
        Instruction-Mining & \ding{56} & \ding{52} & MT-Bench & OPENLLM, MT-Bench \\
        CaR & \ding{52} & \ding{52} & Vicuna, PandaLM, CoachLM, Self-instruct & - \\
        SelectIT & \ding{52} & \ding{56} & - & MMLU, TYDIQA, BBH, GSM, HumanEval, AlpacaEval \\ 
        InstructionGPT-4 & \ding{52} & \ding{56} & LLaVA-Bench & MME, VQA, MMBench \\
        IFD & \ding{52} & \ding{56} & Vicuna, Koala, WizardLM, Self-Instruct, LIMA & OPENLLM \\ 
        Superfiltering & \ding{52} & \ding{56} & WizardLM & OPENLLM, AlpacaEval \\
        Nuggets & \ding{52} & \ding{56} & - & MT-Bench, AlpacaEval, MMLU, HumanEval \\  
        LESS & \ding{52} & \ding{56} & - & MMLU, TYDIQA, BBH \\
        SHED & \ding{52} & \ding{56} & - & MMLU, ARC-challenge \\ 
        MoDS & \ding{52} & \ding{56} & Koala, WizardLM, Self-instruct, Vicuna, LIMA & - \\ 
        InsTag & \ding{56} & \ding{52} & - & MT-Bench \\
        DEITA & \ding{56} & \ding{52} & - & OPENLLM, MT-Bench, AlpacaEval \\ 
        \hline
    \end{tabularx}
    \caption{The counterpart models and evaluation metrics used in each method.}
    \label{tab:DataSum2}
\end{table*}

\textbf{Community Model Preference.} DEITA~\cite{liu2024makesgooddataalignment} adopts the Evol-Instruct method~\cite{xu2023wizardlm} to generate samples with varying complexity and quality, using them to train community models (e.g., LLaMA) as quality scorers instead of ChatGPT. The model separately assesses instruction complexity and response quality scores for each candidate sample. A score-first, diversity-aware algorithm is then designed to rank and select samples based on a combined evaluation of these scores. To optimize selection cost, CaR~\cite{ge2024clustering} introduces an expert-aligned, diversity-preserving approach by training a lightweight Sentence-BERT model (135M parameters) as the quality scorer. It then applies clustering and ranking techniques to select top-K candidates while maintaining diversity.

\subsubsection{Sample's Influence on Model Output as Continuous Quality Labels} 
\label{sec:SampleInfluence}

Another line of research calculates the sample's influence on the model's final performance as a continuous quality label for data selection.

\begin{figure*}[!t]
\centering
\includegraphics[width=0.85\textwidth]{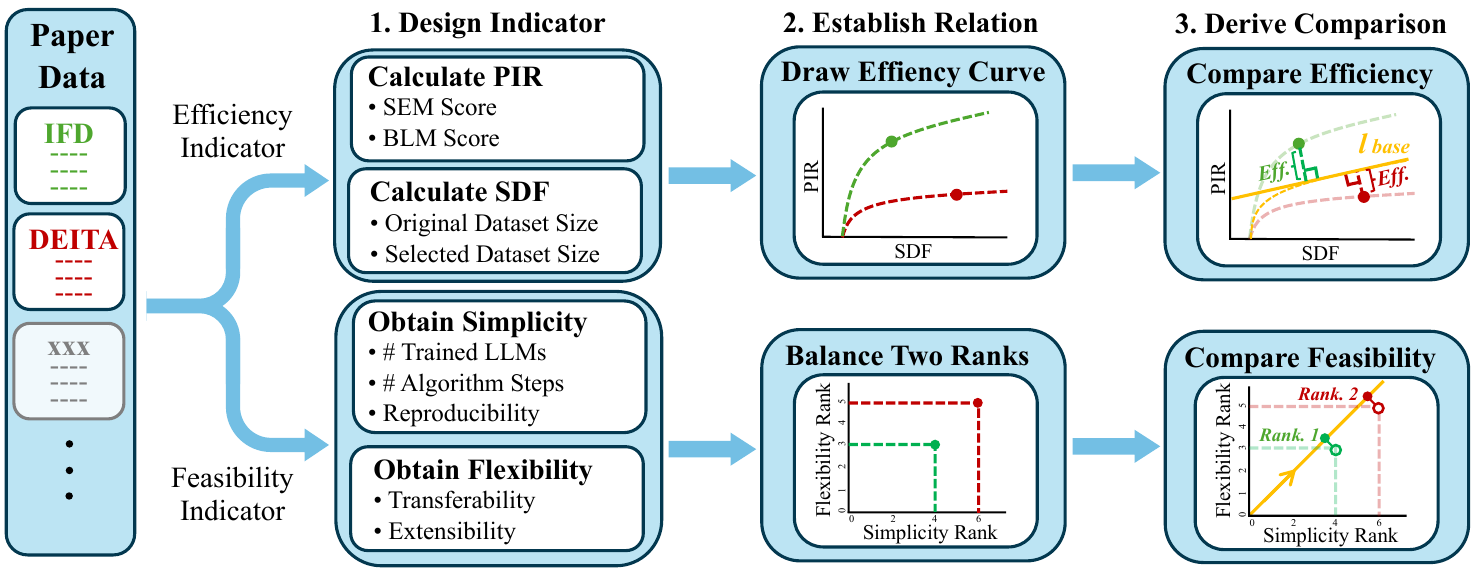}
\caption{
The unified comparison framework, in which IFD and DEITA are chosen for clarity and ease of understanding. 
The design indicator stage calculates statistics based on papers' reported data, such as Performance Improvement Ratio (PIR) and Selected Dataset Fraction (SDF). 
The establish relation stage visualizes the interplay between these indicators through graphs. 
Finally, the derive comparison stage draws conclusions from the graphs.
}
\label{fig:Sec method_graph}
\end{figure*}

Most methods leverage model-based estimations, such as inference loss, Shapley values, performance change, or necessity scores, to measure a sample's impact. Instruction-Mining~\cite{cao2023instruction} and InstructionGPT-4~\cite{wei2023instructiongpt4} construct a predictor that utilizes various feature representations of samples to estimate the inference loss of a fine-tuned model, thereby obtaining quality labels. They then employ BLENDSEARCH~\cite{wang2021economic}, which combines global and local optimizations, to determine the optimal dataset size.

To more precisely quantify sample influence at a lower computational cost, SHED~\cite{he2024shed} first applies a model-agnostic clustering algorithm (K-means with Sentence-BERT embeddings) to identify representative proxy samples. It then computes Shapley values for each cluster, using them as quality scores to guide data selection. Similarly, Nuggets~\cite{li2024oneshot} selects samples based on conditional loss, which measures their impact by comparing the model’s performance on a predefined task with and without the sample in context (one-shot score minus zero-shot score). 

MoDS~\cite{du2023mods} follows a two-stage approach: it first selects a high-quality data subset using external model scoring and fine-tunes an initial LLM on this seed dataset. The necessity scores of the samples within this high-quality subset are then used to perform the final data selection. This dual measurement strategy enables MoDS to identify superior samples, leading to significant performance improvements in fine-tuned models.

Unlike these methods, which rely on model predictions, LESS~\cite{xia2024less} assesses a sample's influence by measuring gradient similarity between candidate samples and existing task datasets. It first fine-tunes a LoRA model to extract gradient features from candidate samples, which are then randomly projected into a storage space. Validation samples in the few-shot setting are also projected into this space to compute gradient similarity, which serves as a quality label for data selection.

\section{Selector Evaluation} \label{sec:eval}

To evaluate a selector's effectiveness, existing works compare a Selective-Enhanced Model (SEM), trained on a selected subset of the candidate dataset, against the same foundation model fine-tuned on the full dataset (BaseLine Model, BLM) or other popular LLMs using win-tie-loss metrics or benchmark scores. Table~\ref{tab:DataSum1} and \ref{tab:DataSum2} summarize the evaluation setup, including candidate datasets, comparison models, and evaluation metrics.

\textbf{Candidate Datasets.} Most studies~\cite{li2024superfiltering, li2024quantity, liu2024makesgooddataalignment} use open-source datasets as candidate sources for fine-tuning, aiming to improve performance by selecting higher-quality samples. These datasets can be broadly categorized into two types: (1) Typical datasets, which serve as common instruction-tuning benchmarks, such as Alpaca~\cite{taori2023Alpaca}, Dolly~\cite{conover2023free}, and FLAN~\cite{weifinetuned}; (2) Advanced datasets, which are refinements of typical datasets to improve quality further, such as WizardLM~\cite{xu2023wizardlm}, UltraChat~\cite{ding2023enhancing},  etc.

\textbf{Counterpart Models.} To evaluate SEM’s effectiveness, most works compare it against a BLM, quantifying the improvement achieved through data selection. Popular LLaMA-series models~\cite{lu2023instag, chen2024alpagasus} and Mistral models~\cite{liu2024makesgooddataalignment, xia2024less} are commonly used as backbones. Additionally, some studies~\cite{ge2024clustering, liu2024selectit} benchmark SEM against SOTA models (e.g., GPT-4, Claude, LLaMA-Chat 7B) to assess absolute performance, further validating the selector’s effectiveness.

\textbf{Evaluation Metrics.} The evaluation metric includes both relative and absolute aspects. Relative metrics, like win-tie-loss ratios judged by GPT-4, reflect SEM's performance compared with its counterpart models. Absolute metrics, instead, utilize either the model's response loss on test sets (MMLU) or GPT-4's score on specific benchmarks to evaluate the SEM performance (MT-Bench).

\section{An Unified Comparison Among Data Selection Method}\label{sec:anal}

To directly compare various existing methods while addressing inconsistencies in their experimental setups, we propose a unified comparison approach that incorporates ratio-based efficiency and ranking-based feasibility. Figure \ref{fig:Sec method_graph} illustrates the three-step framework: (1) Design Indicator: evaluate existing methods using designed indicators under a consistent setting (Alpaca as the candidate dataset, LLaMA-2 7B/13B as the SEM, and win-tie-loss against the BLM as the metric); (2) Establish Relation: plot the points of various methods on a two-dimensional space representing Efficiency and Feasibility comprehensively; (3) Derive Comparison: align the methods with a baseline to enable comparison by translating their relationships into accessible measurements. 

\subsection{Efficiency of the Selector}\label{sec:efficiency}

The efficiency of a data selector reflects its ability to accurately identify high-quality data, contributing to better model performance. As shown in the upper branch of Figure \ref{fig:Sec method_graph}, the process consists of three main steps: (1) We define an efficiency space using two ratio-based indicators: the Performance Improvement Ratio (PIR) and the Selected Dataset Fraction (SDF), which adjust for biases in the experimental settings; (2) We establish an efficiency curve to represent the relationship between these indicators and overall efficiency, using extrapolation to align them; (3) We set a baseline and apply a homeomorphism function to convert each method's latent efficiency into an explicit, signed distance from the baseline, enabling direct comparisons.

\subsubsection{Ratio-based Efficiency Indicators}\label{sec:PIRSDF}

\textbf{Performance Improvement Ratio (PIR)}. The PIR is computed as the average ratio of SEM’s performance score to that of the counterpart model across different evaluation metrics. For methods lacking evaluation data under unified conditions, we estimate the PIR by leveraging the consistency of the selector across various settings. Specifically, we compute average ratios for each method under different evaluation conditions and use the observed linear relationship between these ratios to infer the missing PIR values (refer to Appendix~\ref{app:perf}).

\textbf{Selected Dataset Fraction (SDF)}. The SDF quantifies the effect of data size uniformly, calculated as the ratio of the selected dataset's size to the total size of the original candidate dataset. It ensures proportional representation across datasets, eliminating bias due to varying dataset sizes (e.g., from 3,439 samples~\cite{wei2023instructiongpt4} to 306,044 samples~\cite{lu2023instag}).

\subsubsection{Comparison Based on Efficiency Curve} \label{sec:efficiencycurve}
After plotting existing works' PIR and SDF values, their efficiency remains a latent variable. Although higher PIR and lower SDF are generally indicative of greater efficiency, comparing works with different combinations of these indicators can be challenging. To resolve this, we draw inspiration from scaling laws~\cite{kaplan2020scaling, chung2024scaling} and LIMA~\cite{zhou2024lima}, and introduce the efficiency curve extrapolation assumption. We define two key properties of the unified efficiency curve: (1) The curve follows a logarithmic-like pattern, with a rapid but brief increase before approaching linearity; (2) The slope of a superior efficiency curve is always steeper than that of an inferior one (see Appendix \ref{app:extrapolation} for details). Using these properties, we transform efficiency into an explicit representation on the $SDF \times PIR$ space, as shown in Figure \ref{fig:Sec method_graph}.

\begin{figure}[ht]
    \centering
    \includegraphics[width=\linewidth]{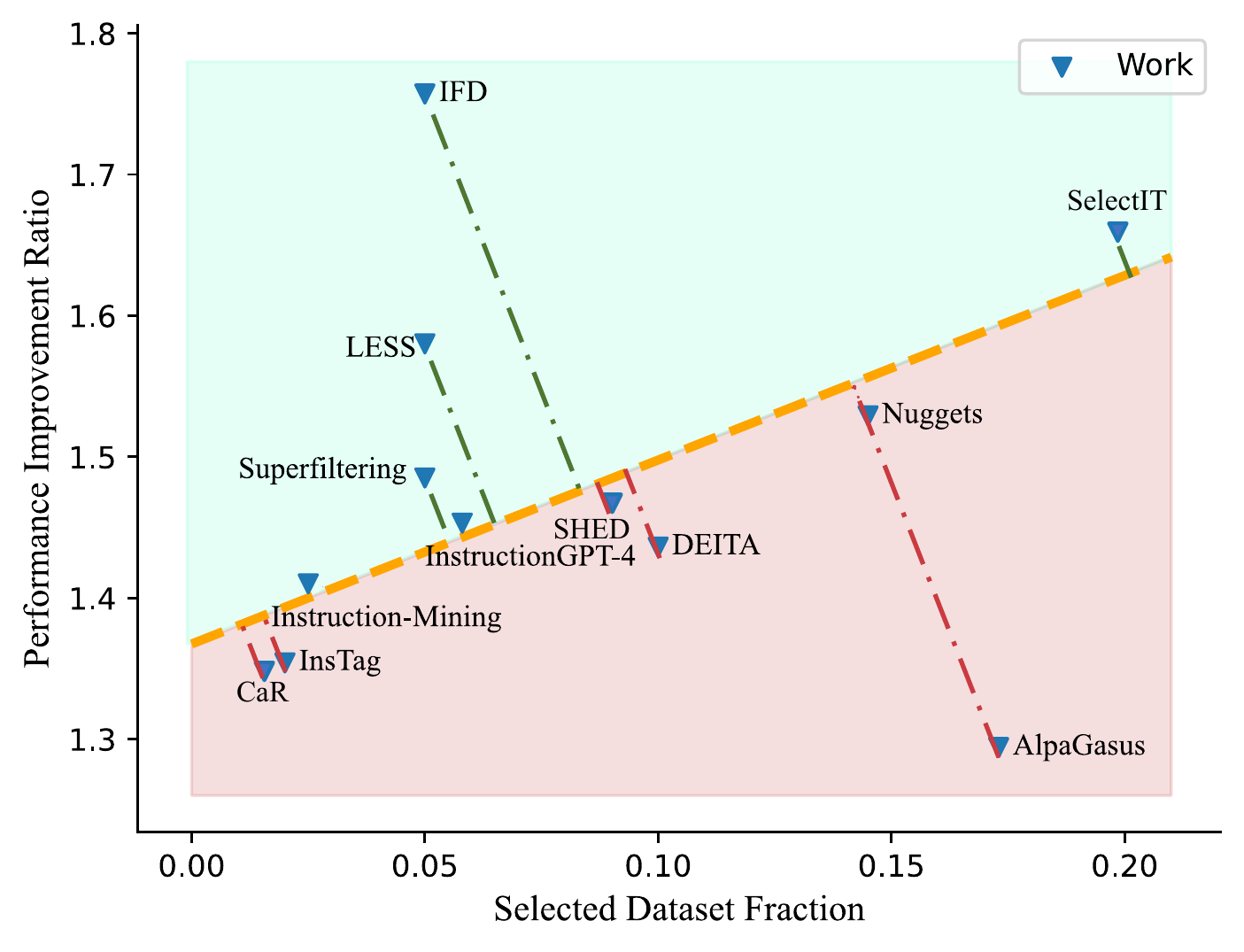}
    \caption{Efficiency comparison of popular data selection methods. The yellow dashed line represents the baseline efficiency $l_{\text{base}}$, with the vertical distance from each method to the baseline indicating its relative efficiency difference.}
    \label{fig:efficiency_compare}
\end{figure}

Additionally, to provide a more direct and robust comparison, we apply a homeomorphism function, transforming relative efficiency comparisons into measurable distances between different methods. Each method’s distance from the baseline represents its comparative efficiency, as shown in Figure~\ref{fig:efficiency_compare}.

Specifically, we take the yellow line connecting Instruction-Mining and Instruction-GPT4 as the baseline, approximating their shared efficiency curve due to their similar selection methods. The signed distance from this baseline indicates whether a method's efficiency is superior (green dashed line) or inferior (red dashed line) to the baseline. Larger distances reflect more significant efficiency differences (see Appendix~\ref{app:comparison}).

As depicted in Figure~\ref{fig:efficiency_compare}, IFD emerges as the most efficient method, while AlpaGasus is the least efficient. This outcome arises because IFD is more targeted: its Instruction Following Difficulty (IFD) score is calculated not only from within the candidate dataset but also using feature extraction from the backbone model. Additionally, its quality labels are derived from the model's loss, which is more direct and avoids reliance on external information. In contrast, AlpaGasus relies solely on ChatGPT scoring without considering the data quality distribution, Backbone characteristics, or optimization objectives, which affects the fine-tuned model’s performance.

Moreover, this comparison enables us to observe the relative rankings of other methods that were previously difficult to compare due to different experimental setups. For example, LESS outperforms Superfiltering, Superfiltering surpasses Nuggets, Nuggets outperform InsTag, and InsTag is better than DEITA.

\subsubsection{Robustness of the Efficiency Indicator} \label{sec:robust}
To mitigate potential bias in the efficiency indicator due to using the same candidate dataset (Alpaca) and backbone model (LLaMA-2) across all works, we conducted additional data selection experiments with two representative methods (AlpaGasus and IFD) using an alternative backbone (LLaMA-3). We obtained their PIR values for comparison.

\begin{table}[!ht]
    \centering
    \resizebox{\columnwidth}{!}{
    \begin{tabular}{lcc}
        \hline
        \textbf{Work} & \textbf{PIR (LLaMA-3 8B)} & \textbf{PIR (LLaMA-2 7B)} \\
        \hline
        AlpaGasus & 1.287 (0.173) & 1.284 (0.173) \\
        IFD       & 1.512 (0.050) & 1.747 (0.050) \\
        \hline
    \end{tabular}
    }
    \caption{Comparison of PIR values between IFD and AlpaGasus with different backbones (LLaMA-2/LLaMA-3) under MT-bench. The value 1.287 (0.173) means the PIR of the AlpaGasus method is 1.512 with an SDF of 0.173. Approaches to obtain the indicators's value are discussed in Section~\ref{sec:PIRSDF}.}
    \label{tab:PIR_comparison}
\end{table}

Tabel~\ref{tab:PIR_comparison} shows that:
(1) IFD consistently outperforms AlpaGasus in efficiency, due to its more targeted design for quality labels, as discussed in Sections~\ref{sec:internal} and~\ref{sec:efficiencycurve};
(2) While AlpaGasus underperforms IFD, it exhibits more stable performance across different backbones, as it is independent of the backbone model when evaluating data quality.

\subsection{Feasibility of the Selector} \label{sec:feasibility}

Unlike previous studies that primarily focus on efficiency, we also take into account the practical usability of each method, a factor often overlooked. To assess this, we follow the process illustrated in the lower branch of Figure~\ref{fig:Sec method_graph} for manual evaluation, as relevant metrics are lacking. (1) We identify five key factors that contribute to two ranking-based feasibility indicators, which capture the simplicity and flexibility of each method, and manually annotate them. (2) We obtain simplicity and flexibility ranks and position the methods on a two-dimensional feasibility measurement plane for relational visualization. (3) We integrate both ranks to derive an overall comparative ranking.

\subsubsection{Ranking-based Feasibility Indicators} \label{sec:SimpFlex}

\begin{figure}[!t]
    \centering
    \includegraphics[width=0.9\linewidth]{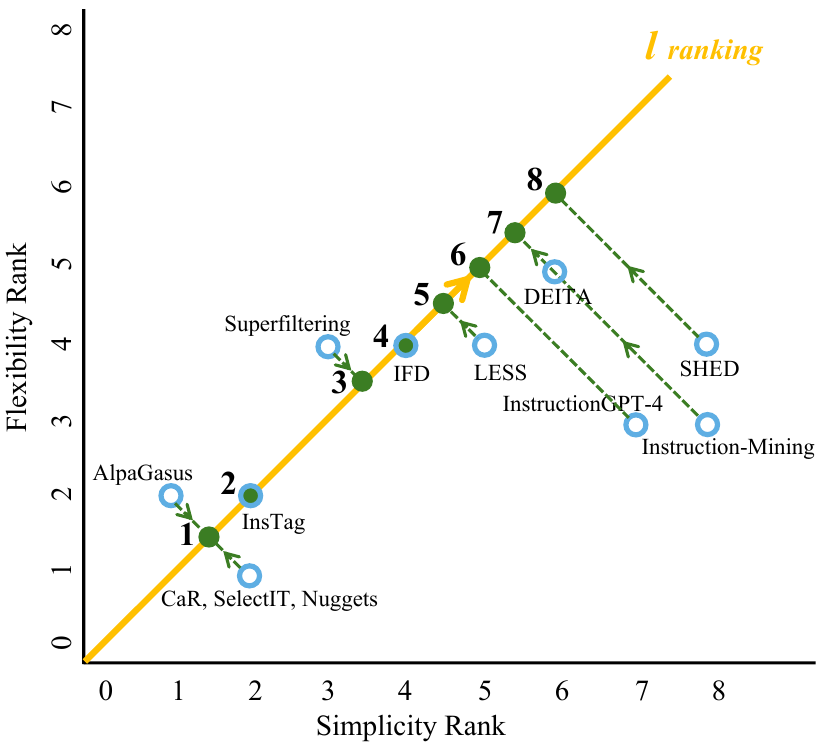}
    \caption{Simplicity and flexibility rank of data selection methods. The yellow line $l_{\text{ranking}}$ is the ranking line, with the projection of each method onto the line indicating its feasibility rank.}
    \label{fig:feasibility_graph}
\end{figure}

\textbf{Simplicity.} This metric assesses the complexity of the selection process and the reproducibility (Rep.) of the method. In terms of complexity, we examine the implementation cost of the method, which includes the number of models trained and the complexity of the executed algorithms. We also verify reproducibility by checking whether the implementation has been open-sourced, as this is highly valued in the research community. Further details can be found in Appendix~\ref{app:simplictiy}.

\textbf{Flexibility.} Additionally, we evaluate the model's flexibility in terms of transferability and extensibility, both critical for a method's generalization in practical applications. 

For transferability, a model-free method allows the selection of different models without retraining the selector, as it does not rely on specific models to obtain quality labels; meanwhile, a dataset-free selector can be applied to any candidate dataset without retraining, highlighting its independence from specific dataset information.

For extensibility, we consider whether the algorithm contains fixed components that limit its adaptability to other algorithms. For instance, methods like AlpaGasus and InsTag rely heavily on third-party tools such as ChatGPT, which hinders their ability to extend to other frameworks due to the immutability of ChatGPT.

\subsubsection{Comparison with Simplicity and Flexibility} \label{sec:feasibility_compare}

Based on the published papers for each method, we annotate their simplicity and flexibility and plot them to derive the final rankings, as shown in Figure~\ref{fig:feasibility_graph}. We assign equal weight to simplicity and flexibility and draw the $l_{\text{ranking}}$ line to provide a comprehensive ranking based on both factors. Specifically, we project each method perpendicularly onto the ranking line (which passes through the origin), representing the average of the two ranks. The sequence of the projected points corresponds to the feasibility ranking of the methods.

As shown in Figure~\ref{fig:feasibility_graph}, AlpaGasus demonstrates the highest flexibility, despite its mediocre performance. Its simplicity, stemming from not requiring LLM training and involving fewer steps in the selection process, makes it easy for others to reproduce its results—even though no official implementation has been released. Moreover, it can be more easily transferred to other scenarios (both model-free and dataset-free) as it relies solely on ChatGPT scoring without scenario-specific information. Although Instruction-Mining and SHED outperform AlpaGasus, they compromise feasibility due to their heavy reliance on fine-tuning multiple LLMs and using complex quality indicators derived from both datasets and models.

\subsection{Overall Consideration of the Selector} \label{sec:overall}
In summary, a key observation from our comparison of efficiency and feasibility is that existing methods often struggle to achieve both high performance and ease of use simultaneously. Additionally, we find that the more tailored a data selection method is, the better the SEM performance tends to be. For instance, DEITA is more complex than AlpaGasus because it trains the selector based on a backbone model and considers data diversity, leading to better performance. However, these more complex processes and algorithms can introduce external factors that hinder direct optimization and reduce transferability. For example, despite its tailored approach, LESS underperforms IFD due to the added complexity and noise introduced by external datasets.

We also evaluate MoDS, which demonstrates high comparative efficiency (PIR = 3.34, SDF = 0.02), but it uses two coupled phases of LLM training, similar to applying the selection method twice, resulting in its exclusion from comparisons with other works in our paper.

\section{Discussions} \label{sec:disc}

\subsection{Trends} \label{sec:trend}

\begin{figure}[ht]
    \centering
    \includegraphics[width=\linewidth, page=1]{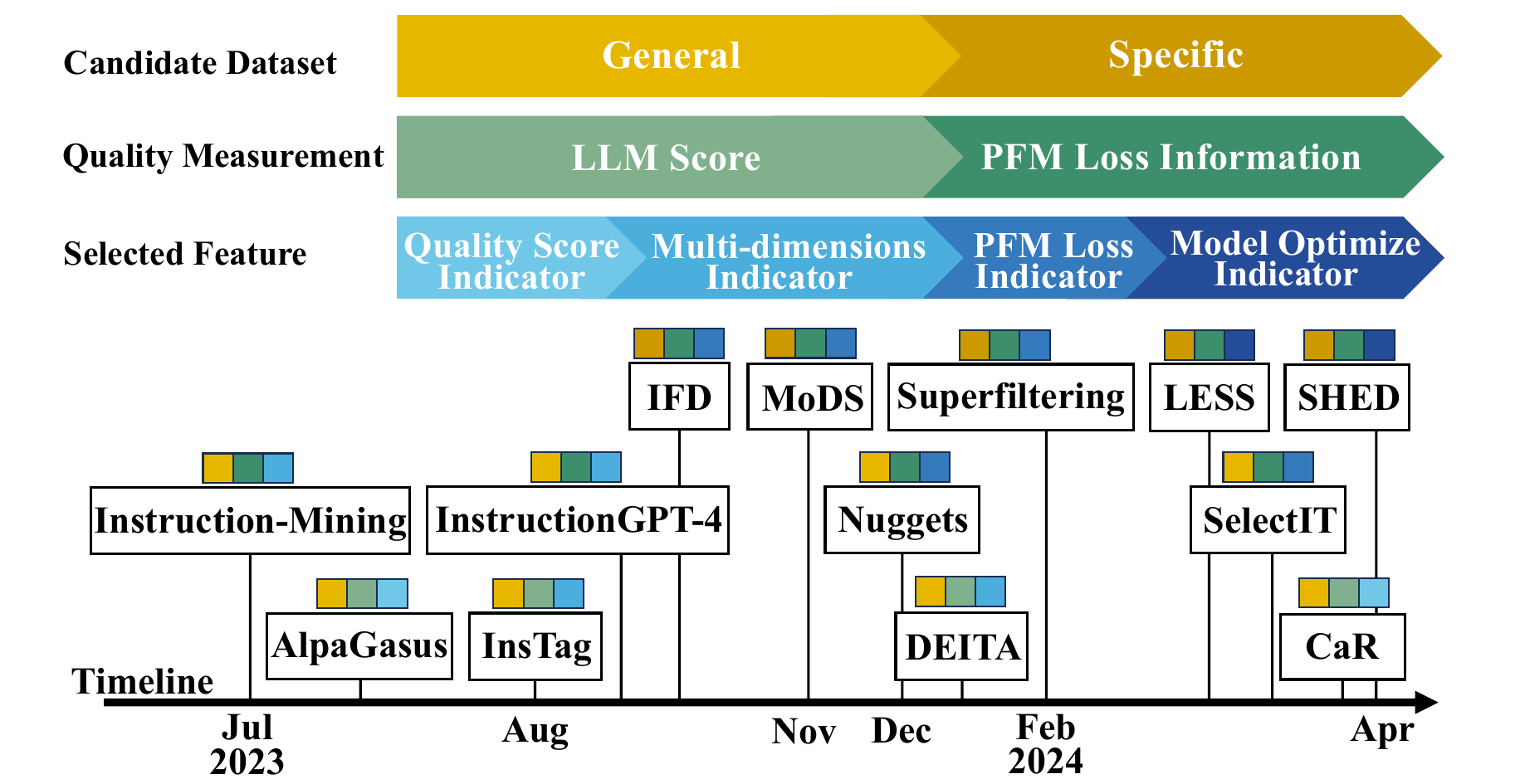}
    \caption{The timeline of the evolution of data selection methods.}
    \label{fig:Sec7_T1}
\end{figure}

Figure~\ref{fig:Sec7_T1} illustrates the current research trends, where we have categorized the existing work following three key aspects: Candidate Datasets, Quality Measurement, and Selected Features, arranged chronologically. Based on this, we can identify three main trends:

(1) Candidate Datasets: As detailed in Section \ref{sec:eval}, earlier works predominantly utilized general datasets for constructing selectors~\cite{cao2023instruction}. However, newer studies have shifted towards using more specific datasets tailored for selector development~\cite{li2024quantity}.

(2) Quality Measurement: There has been a noticeable shift in how quality is measured, as mentioned in Section \ref{sec:selector}. Early approaches focused on external scorer-based judgments that assessed sample quality and diversity~\cite{chen2024alpagasus, liu2024makesgooddataalignment}. In contrast, recent methods have moved towards direct measurement of model performance improvement based on the influence of individual samples~\cite{xia2024less}.

(3) Selected Features: As discussed in Sections \ref{sec:preprocess} and \ref{sec:selector}, they are shifted from using concrete indicators (single quality score~\cite{chen2024alpagasus} and multi-dimension indicators~\cite{lu2023instag}) to abstract indicators (the training loss~\cite{li2024superfiltering} and gradient similarity~\cite{xia2024less}).

\subsection{Challenges} \label{sec:challenge}
Despite substantial progress in data selection for fine-tuning large language models (LLMs), several critical challenges remain, particularly in addressing the open question: \textbf{How can we design effective sample quality measurements for data selection?}

\textbf{(1) Balancing the Efficiency and Feasibility}. As discussed in Section~\ref{sec:anal}, more targeted approaches that directly measure the sample's impact on model performance often demonstrate higher efficiency. However, these methods can reduce feasibility when applied to diverse settings, a limitation that has been overlooked in previous studies~\cite{cao2023instruction, wei2023instructiongpt4, xia2024less}. Developing a measurement strategy and data selection algorithm that effectively balances both efficiency and feasibility remains a challenging research goal.

\textbf{(2) Ensuring the Measurement Objectivity}. For methods that rely on existing LLM preferences~\cite{chen2024alpagasus, lu2023instag, liu2024makesgooddataalignment}, designing appropriate scoring prompts to ensure reliable scores is crucial. This is particularly challenging given that biases persist even in state-of-the-art models such as GPT-4/GPT-4o. For methods based on a sample's influence~\cite{li2024oneshot, li2024quantity, xia2024less}, creating metric functions that account for both sample characteristics and model specifics is essential. Therefore, achieving objective and consistent measurements across different tasks and models remains a significant challenge.

\textbf{(3) Improving Specific Tasks/Domains Performance without Compromising Others}. Existing research~\cite{jha2023limit} shows that performance improvements from selected samples vary across tasks. While there are notable gains in tasks like writing and role-playing, improvements in mathematics and reasoning are often marginal~\cite{li2024oneshot,lu2023instag}. Therefore, developing methods to improve performance in task-specific domains without negatively affecting other tasks is an important direction for future research.
 
\textbf{(4) Satisfying Multiple Goals in Data Selection}. Most existing works focus on merging various aspects such as data quality~\cite{chen2024alpagasus, xia2024less} and diversity~\cite{lu2023instag, liu2024makesgooddataalignment}. However, data selection methods for multi-turn and multi-model scenarios are still underdeveloped. Developing approaches that address these complex scenarios is crucial for expanding the applicability of data selection techniques.

\section{Conclusion} \label{sec:conclusion}
In this paper, we present a fine-grained survey of a dozen data selection methods for fine-tuning large-scale language models and establish a three-stage data selection scheme to standardize the process. To address inconsistencies across different experimental setups, we also introduce a unified comparison approach using ratio-based and ranking-based indicators to quantify efficiency and assess feasibility, which has been overlooked in previous works. Our comprehensive evaluations, considering both efficiency and feasibility, reveal that methods with more targeted designs tend to achieve higher efficiency, though often at the cost of feasibility. Additionally, we outline the progression of existing research and identify four critical challenges for future work:  balancing efficiency and flexibility, ensuring measurement objectivity, improving specific tasks/domain performance without compromising others, and satisfying multiple goals in data selection. These challenges point to key areas for future research, particularly in refining sample quality measurement techniques.

\section*{Limitation}
We mainly research data selection for instruction fine-tuning LLMs instead of data rewriting or augmentation. Although we have already comprehensively examined the existing works, we acknowledge that there may still be some works we neglected, especially the very recent work that was published on the preprint platforms.

Besides, we focus on outlining the scheme of existing work on data selection and propose an analytical method for comparing various works directly. Therefore, the descriptions of each work could be limited to key points relevant to our study rather than providing a comprehensive overview due to limited space.

\section*{Acknowledgements}
This research is supported by the project of Shenzhen Science and Technology Research Fund (Fundamental Research Key Project Grant No. JCYJ20220818103001002), Shenzhen Science and Technology Program (Grant No. ZDSYS20230626091302006), Key Project of Shenzhen Higher Education Stability Support Program (Grant No. 2024SC0009) and SRIBD Innovation Fund (Grant No. K00120240006).


\bibliography{custom}
\appendix

\newpage

\section{Appendix}
\label{sec:appendix}

\subsection{Performance Improvement Ratio} \label{app:perf}

\begin{table}[!ht]
     \centering
     \resizebox{\linewidth}{!}{
     \begin{tabular}{p{0.42\linewidth}p{0.31\linewidth}p{0.12\linewidth}p{0.15\linewidth}p{0.12\linewidth}p{0.15\linewidth}}
     \hline
         \textbf{Method} & \textbf{Backbones of SEM} & \multicolumn{2}{c} {\textbf{Same Model}} & \multicolumn{2}{c}{\textbf{Other Models}}\\ \cline{3-4} \cline{5-6}
         ~ & ~ & \textbf{Win Rate} & \textbf{Bench.} & \textbf{Win Rate} & \textbf{Bench.}\\
         \hline
         \textbf{AlpaGasus} & LLaMA-2-7B & 1.284 & 0.949 & -& - \\ 
         \textbf{Superfiltering} & LLaMA-2-7B & 1.475 & 1.010 & - & - \\ 
         \textbf{InsTag} & LLaMA-2-13B & 1.344 & - & - & 0.985   \\ 
         \textbf{DEITA} & LLaMA-2-13B & 1.426 & -  & - & 1.000 \\ 
         \textbf{InstructionGPT-4} & MiniGPT-4 & 1.443 & - & - & - \\ 
         \textbf{Nuggets} & LLaMA-2-7B & 1.519 & - & - & -  \\
         \textbf{IFD} & LLaMA-2-7B & 1.747 & - & - & - \\ 
         \textbf{LESS} & LLaMA-2-13B & (1.491) & 1.015  & - & - \\ 
         \textbf{Instruction-Mining} & LLaMA-2-7B & (1.400) & - & 0.212 & 0.991 \\ 
         \textbf{SHED} & LLaMA-1-7B & (1.460) & 1.005 & - & - \\
         \textbf{CaR} & LLaMA-2-7B & 1.343 & - & - & - \\
         \textbf{SelectIT} & LLaMA-2-7B & (1.653) & 1.067& - & - \\

         \hline
     \end{tabular}
     }
     \caption{ The performance improvement under four evaluation settings. In the Same Model, we compare SEM and BLM, while in other models, we compare SEM and the same-size models trained based on other backbones (such as LLaMA Chat).}
     \label{tab:PIR}
\end{table}
Since different work uses different evaluation methods, it is difficult to compare them directly. Therefore, to uniformly evaluate their performance, we first fix Alpaca as the candidate dataset since it is widely adopted, and then we divide the various evaluation settings mentioned in all works into (1) BLM comparing with SEM on the same Backbone and (2) BLM comparing with other models (such as LLaMA Chat). Then, we further divide them into win rate and benchmark improvement (Bench.) with different kinds of evaluation metrics.

In total, we have four evaluation settings, as shown in Table~\ref{tab:PIR}, and we take the average of each type in Formula (\ref{formula:A1}) if it uses multiple evaluation metrics or other models. 
\begin{equation}
    PIR = \frac{1}{n} \sum_{i=0}^n \frac{X_i}{Y_i}
    \label{formula:A1}
\end{equation} where $X_i$ and $Y_i$ are, respectively, the performance of the SEM and the counterpart model under the same evaluation setting $i$, and $n$ is the total number of the evaluation settings using the same kind of evaluation metric (win-tie-loss or benchmark scoring) and counterpart model. We then choose the win rate under BLM as the ratio indicator of PIR not only because it directly reflects the improvement effect made by the selector but also because most of the works provide this value.

To fill the missing value, we leverage the consistency of model performance: the same model should perform similarly under different evaluation settings. Therefore, we obtain the bridge function by linearly regressing the other works with the win rate under BLM as the label and the other three entries in Table \ref{tab:PIR} as variables. Then, we estimate the missing value by using the bridge function to transfer the value under other entries into PIR. 

For the work that does not adopt the Alpaca as a candidate dataset, we use the same regression method to scale its performance to the Alpaca dataset. Moreover, for the work evaluated under multiple SDF and thus having multiple PIR, we choose the one that has the highest PIR value to indicate the method's optimal performance.
\begin{figure}[ht]
    \centering
    \includegraphics[width=\linewidth]{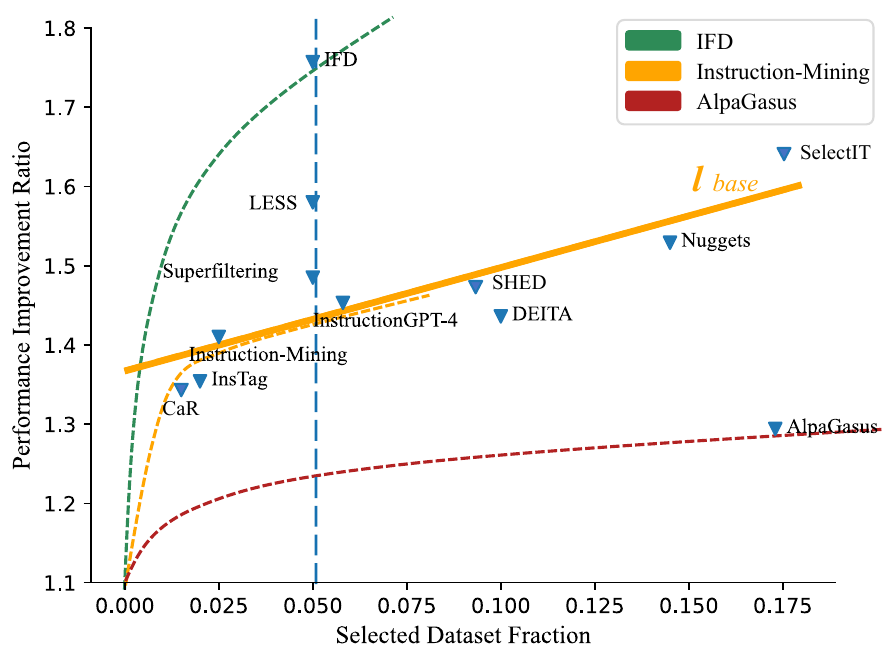}
    \caption{Efficiency curves of three representative methods (IFD, Instruction-Mining, and AlpaGasus). The yellow baseline connects Instruction-Mining and InstructionGPT-4, serving as the shared asymptote of their efficiency curves and providing a ground approximation for efficiency trends.}
    \label{fig:efficiency_curve}
\end{figure}

\subsection{Efficiency Curve Extrapolation} \label{app:extrapolation}

To directly compare the efficiency of work with different performance improvement ratios (PIR) and selected data fractions (SDF), we establish the efficiency curve for each work using the efficiency curve extrapolation, which consists of two properties:

\textbf{Property 1}.\label{property_1} Relevant theories (such as scaling law)~\cite{kaplan2020scaling, sun2017revisiting, moskovskaya2023predicting} suggest that the loss value is linear with respect to the log of data size given that the dataset always contains the same portion of good and bad data. Inspired by these theories, we model the PIR-SDF relation of a "fixed-quality" dataset to be a log-like curve, which is increasing, concave down, and approaching linear after an initial increase.

\textbf{Property 2}.\label{property_2} The efficiency represented by a higher curve with a larger slope is superior to that represented by a lower one. It is intuitively derived from property 1 and the fact that high-quality data leads to better SEM performance \cite {zhou2024lima}. Therefore, if a method is more efficient, then the portion of good data it selects will always be larger, and thus the gradient of its performance improvement ratio (PIR) will always be greater than an inferior method at any given selected dataset fraction (SDF) in the graph.

Figure \ref{fig:efficiency_curve} is plotted following the efficiency curve extrapolation. The curves are represented with dashed lines to show that they are not ground-truth curves but rather extrapolations.

\subsection{Efficiency Comparison Method}\label{app:comparison}
The extrapolation idea introduced in Appendix \ref{app:extrapolation} is an intuitive tool for efficiency comparison because it allows different data selection methods to slide on their respective efficiency curve, scaling to any SDF value with a matched PIR while maintaining their unique efficiency trait. For example, all works in Figure~\ref{fig:efficiency_curve} can slide to the blue line following their unique efficiency curves and then get their rankings directly according to the corresponding PIR value since they all have the same SDF=0.050.

However, this naive comparison method faces two challenges: (1) except for the yellow baseline, all other works' ground truth efficiency curve is not accessible due to the lack of data; (2) such a method gives only an intuitive ranking rather than concrete values, which are necessary for more nuanced comparison.



To address these two challenges, we propose to use each work's distance to the baseline to represent their efficiency value as illustrated in Figure~\ref{fig:feasibility_graph}, which is supported by the idea of homomorphism. Related math derivation is listed below.

We start by formally defining this problem. Let $X$ denote the latent efficiency space, where each $x\in X$ represents the intrinsic efficiency of a data selection method. Our goal is to compare $x_k$ (efficiency of method $k$) with $x_b$ (efficiency of the baseline method) using their observable operating points in the product space $S\times P$, where $S$ is the selected dataset fraction (SDF) and $P$ is the performance improvement ratio (PIR).

In order to establish a one-to-one correspondence between latent efficiency and observable data, we formulate the efficiency curve as a homomorphism by the following definition: the efficiency curve of method $k$ is a function $f_k:S \to P$, parameterized by $x_k$, mapping SDF to PIR. Formally,

\begin{equation}
f_k(s) = f(s;x_k),
\end{equation}
where $f$ is a homeomorphism between $X$ and $S\times P$. This definition captures the bijective and continuous property of the efficiency curve: (1) \textit{Bijectivity}: Each $x_k \in X$ maps to a unique curve $f_k(s)$, and vice versa; (2) \textit{Continuity}: If $x_1 \succ x_2$ (higher efficiency), then $(f(s; x_1) > f(s; x_2)$ for all $s \in S$.  

Following the definition, we can now quantify the \textbf{property 2} (mentioned in Appendix \ref{app:extrapolation}) of an efficiency curve by representing the efficiency differences via slope accumulations. We argue for this idea by first interpreting the meaning of the slope: The derivative $\frac{df_k}{ds}$ reflects how efficiently a method trades off SDF for PIR (aka. how efficiently it can select good data). Formally, for $x_1 \succ x_2$,

\begin{equation}
\frac{df(s; x_1)}{ds} > \frac{df(s; x_2)}{ds} \quad \forall s \in S
\end{equation}
Steeper slopes indicate better efficiency (greater PIR gain per unit SDF increase). 

Now, with the per SDF efficiency advantage gauged by the slope difference, we can further quantify the accumulative total efficiency difference between method $k$ and baseline $b$ over $S$ by defining their vertical separation as:

\begin{equation}
f_k(s) - f_b(s) = \int_{s_{\text{min}}}^s \left( \frac{df(\tau; x_k)}{d\tau} - \frac{df(\tau; x_b)}{d\tau} \right) d\tau
\end{equation} \label{eq:efficiency_difference}
Equation (4) integrates slope differences across $S$, enabling a quantitative description of how much better one method is over the baseline. 

Since future comparisons will mostly be based on the baseline, a simplified notation is derived as follows: Let $f_b(s) = f(s; x_b)$ be the baseline’s known efficiency curve. For method $k$ operating at $(s, k)$, its vertical offset from the baseline at $s_k$ is:

\begin{equation}
\Delta p_k = p_k - f_b(s_k)
\end{equation}
\(\Delta p_k\) measures how much \(k\) outperforms the baseline at the same SDF.

To further map the efficiency to directly observable data in $S\times P$, we apply geometric correction by transforming the vertical offsets $\Delta p_k$ into a perpendicular distance $y_k$ (see Figure~\ref{fig:feasibility_graph}). Formally, let $\theta$ be the angle between the baseline curve and the SDF axis at $s_k$. We define $y_k$ as:  
\begin{equation}
y_k = \Delta p_k \cdot \sin\theta
\end{equation}

We end this derivation by giving the final mapping function and argue that it's still a homeomorphism, which enables a neat and rigorous comparison. Define $\varphi: X \to \mathbb{R}$ as
\begin{equation}
\varphi(x_k) =  \sin\theta \cdot \left( f(s_k; x_k) - f(s_k; x_b) \right)
\end{equation}
We argue that $\varphi$ is a homomorphism by checking whether its bijective and continuous: (1) Bijectivity: Guaranteed by the bijectivity of $f$; (2) Continuity: Follows from the continuity of $f$ and $sin(\theta)$. Thus, we claim that this final mapping function preserves the ranking $x_1 \succ x_2 \iff y_1 > y_2$. 

In conclusion, for any method $k$ with observable $(s_k, p_k)$, its efficiency distance to the baseline is:

\begin{equation}
y_k = \sin\theta \cdot \left( p_k - f_b(s_k) \right),
\end{equation}
equivalent as Equation (7), requiring only a single point and thus bypassing the need for the inaccessible efficiency curve.

\begin{table*}[!ht]
    \centering
    \resizebox{\linewidth}{!}{
    \begin{tabular}{p{0.2\linewidth}p{0.18\linewidth}p{0.2\linewidth}p{0.08\linewidth}p{0.13\linewidth}p{0.15\linewidth}p{0.12\linewidth}p{0.23\linewidth}p{0.13\linewidth}p{0.13\linewidth}}
        \hline
        \textbf{Methods} & \# Trained LLMs & \# Algorithm Steps & Rep. & \textbf{Simplicity}  & \multicolumn{2}{c}{Transferability} & \multicolumn{1}{c}{Extensibility} & \textbf{Flexibility} & \textbf{Feasibility}\\ \cline{6-8} 
         ~  &  ~ & (\# Using LLMs) & ~ & ~ & Model Free & Dataset Free & ChatGPT/GPT-4 Free\\\hline
        AlpaGasus & 0 & 2(1) & \ding{56} & \textbf{1} & \ding{52} & \ding{52} & \ding{56} & \textbf{2} & \textbf{1}\\
        SelectIT & 0 & 4(3) & \ding{52} & \textbf{2} & \ding{52} & \ding{52} & \ding{52} & \textbf{1} & \textbf{1}\\
        Nuggets & 0 & 4(2) & \ding{52} & \textbf{2} & \ding{52} & \ding{52} & \ding{52} & \textbf{1} & \textbf{1}\\
        Car & 0 & 4(0) & \ding{52} & \textbf{2} & \ding{52} & \ding{52} & \ding{52} & \textbf{1} & \textbf{1}\\
        InsTag & 0 & 3(1) & \ding{56} & \textbf{2} & \ding{52} & \ding{52} & \ding{56} & \textbf{2} & \textbf{2} \\
        Superfiltering & 1* & 3(1*) & \ding{52} & \textbf{3} & \ding{56} & \ding{56} & \ding{52} & \textbf{4} & \textbf{3}\\
        IFD & 1 & 3(1) & \ding{52} & \textbf{4} & \ding{56} & \ding{56} & \ding{52} & \textbf{4} & \textbf{4}\\
        LESS & 1 & 4(2) & \ding{52} & \textbf{5} & \ding{56} & \ding{56} & \ding{52} & \textbf{4} & \textbf{5}\\
        InstructionGPT-4 & 30 & 4(1) & \ding{52} & \textbf{7} & \ding{56} & \ding{52} & \ding{56} & \textbf{3} & \textbf{6}\\
        DEITA & 2 & 5(4) & \ding{52} & \textbf{6} & \ding{56} & \ding{56} & \ding{56} & \textbf{5} & \textbf{7} \\
        Instruction-Mining & 129 & 4(0) & \ding{56} &\textbf{8} & \ding{56} & \ding{52} & \ding{56} & \textbf{3} & \textbf{7}\\
        SHED & 500 & 3(0) & \ding{52} &\textbf{8} & \ding{56} & \ding{56} & \ding{52} & \textbf{4} & \textbf{8}\\
         \hline
    \end{tabular}}
    \caption{Feasibility rank considers both Simplicity rank and Flexibility rank. The former consists of three indicators: (1) \# Trained LLMs; (2) \# Algorithm Steps (\# Times Using LLMs in the algorithm) and (3) Reproducibility, while the latter considers extensibility and transferability. The number in the bracket of the "\# Algorithm Steps" column indicates the times of LLMs used in the selection algorithm. * indicates that Superfiltering trains a GPT-2 instead of LLaMA. The detail of each step of the "\# Algorithm Steps" is shown in Algorithm 1-11.}
    \label{tab:Feasibility}
\end{table*}

\subsection{Feasibility}
\label{app:feasibility}
We consider simplicity and flexibility to be the two main aspects when evaluating a selection method's feasibility. This section explains how these two aspects are qualitatively and reasonably evaluated using further refined indicators.

\subsubsection{Simplicity}
\label{app:simplictiy}
The simplicity of a data selection method takes into account (1) the number of LLMs trained in selector construction, (2) the number of steps in the selection algorithm, and (3) reproducibility, which is based on the open-source state of the code.

\textbf{\# of Trained LLMs}. This indicator counts the number of LLMs trained during the selector construction stage. For example, AlpaGasus, InsTag, CaR, and Nuggets use purely ChatGPT (commercial LLM), LLaMA (community model in Nuggets and SelectIT), or expert small model (355M in CaR) as a scorer or tagger, so no LLM is trained in the construction and the count is thus 0. IFD, Superfiltering, and LESS train one warm-up model (LLaMA for IFD and LESS, GPT-2 for Superfiltering) to obtain quality labels for candidate datasets, so the count is 1. MoDS trains one intermediate model for necessity evaluation, so the count is also 1.  DEITA trains a complexity scorer and a quality scorer from ChatGPT-evolved data separately, so the count is 2. Instruction-Mining fine-tunes 129 models to obtain loss scores on 129 data subsets to rule-fit a linear loss score predictor, so the count is 129. The same count rule applies to InstructionGPT-4 since these two works are almost identical in method. Additionally, SHED fine-tunes 500 models during the proxy Shapley value calculating stage, so the count is 500.

\textbf{\# of Algorithm Steps}. The following pseudo algorithms help count the steps in the selecting stage (excluding the Init and Return steps), where the number in the bracket in the table is the number of LLMs used. For example, based on the Algorithm \ref{alg:alpagasus}, AlpaGasus performs first ChatGPT scoring and then ranking to get the final selected subset, which consists of 2 steps with 1 LLM usage. 

\textbf{Reproducibility}. "\ding{52}" means the code is open-source on GitHub, "\ding{56}" means the opposite. Additionally, AlpaGasus has been open-source by others but not by the authors, and InsTag provides a demo on ModelScope and checkpoints on HuggingFace, but no codes are open-source. Thus, we consider them to be close-source.

\subsubsection{Flexibility}
\label{app:flexibility}
The flexibility of a selection method considers both transferability and extensibility. The former is determined by whether the method can be transferred to other methods without retraining for every new model choice or candidate dataset choice, while the latter is by whether proprietary models like ChatGPT/GPT4 are used in the selection process.

\textbf{Model Free}. It means the selector model can be replaced with any other model without harming the efficiency of the data selector conceptually. For example, AlpaGasus allows users to replace the ChatGPT with other models for direct scoring, which has the downside of over-reliance on proprietary models. Similarly, InsTag prompts GPT for annotation; CaR uses an expert small model for expert-aligned scoring; Nuggets prompts LLaMA for conditional loss; and SelectIT obtains three-level of uncertainties directly from a foundation model, which does not need to retrain the selector if the model is changed. Meanwhile, Superfiltering, IFD, and LESS need to warm up a model with a portion of the candidate dataset, and thus, changing the model requires a re-warm-up and is thus not model-free. DEITA trains two separate scorers from LLaMA. Replacing LLaMA with another model requires retraining the scorers with the same seed-evolved data. Instruction-Mining and InstructionGPT-4 need to retrain the two sets of models for the loss estimation; therefore, they are not model-free. SHED performs actual fine-tuning to calculate Shapley values, while MoDS trains an intermediate model for necessity evaluation. Therefore, both are not model-free.

\textbf{Dataset Free}. It means the change of the candidate dataset won't require retraining the selector, which means the candidate dataset is not used as training data in the construction of the data selector. For example, from AlpaGasus to SelectIT in the table, all these works use the candidate dataset directly for score obtaining without training, so they are dataset-free. For instruction-Mining/GPT-4, they derive the mapping function to estimate loss on any candidate dataset. Works like Superfiltering, IFD, LESS, and DEITA all rely on the information within a given candidate dataset and need warm-up; retraining is necessary if the dataset is replaced. SHED calculates Shapley by iteratively fine-tuning the model, which requires re-finetuning if the candidate dataset is changed.



\subsection{Algorithms Summary} \label{app:algorithm}
We summarize the key selection algorithm in a highly abstract way to facilitate the qualitative evaluation of \textbf{Simplicity} indicator, where algorithm steps are counted. The following summary uses a modular way to count algorithm steps, attempting to ensure each step is of similar difficulty while capturing all the necessary steps.
The counting result is shown in Table \ref{tab:Feasibility}, where the init and return steps of each method do not count towards the number of algorithm steps.

\begin{algorithm}
\caption{AlpaGasus}
\label{alg:alpagasus}
\begin{algorithmic}[1]
\STATE \textbf{Init} $D$ = Candidate Dataset, $S$ = ChatGPT, $U$ = LLM Usage
\STATE Use $S$ to score $D$ ($U$+=1)

--> sample with score
\STATE Do score ranking and pick top K
\STATE \textbf{Return} Selected Subset
\end{algorithmic}
\end{algorithm}

\begin{algorithm}
\caption{InsTag}
\label{alg:instag}
\begin{algorithmic}[1]
\STATE \textbf{Init} $D$ = Candidate Dataset, $S$ = ChatGPT, $U$ = LLM Usage
\STATE Use $S$ to tag $D$ ($U$+=1) 

--> sample with tags
\STATE Do tag normalization

--> sample with tag statistics
\STATE Do complexity-first diverse sampling
\STATE \textbf{Return} Selected Subset
\end{algorithmic}
\end{algorithm}

\begin{algorithm}
\caption{CaR}
\label{alg:car}
\begin{algorithmic}[1]
\STATE \textbf{Init} $D$ = Candidate Dataset, $S$ = Expert Model(355M), $U$ = LLM Usage=0
\STATE Use $S$ to score $D$

--> sample with score

\STATE Do score ranking and pick top n1

--> selected dataset $D_{n1}$

\STATE Do clustering on $D$

--> K subsets $D'$
\STATE Do score ranking in each $D'$ and pick top n2 in each $D'$

--> selected dataset $D_{K*n2}$
\STATE \textbf{Return} $D_{n1}+D_{K*n2}$
\end{algorithmic}
\end{algorithm}

\begin{algorithm}
\caption{Nuggets}
\label{alg:nuggets}
\begin{algorithmic}[1]
\STATE \textbf{Init} $D$ = Candidate Dataset, $S$ = Backbone, $U$ = LLM Usage
\STATE Prompt $S$ with zero-shot $D$ ($U$+=1) 

--> sample with ZeroShotScore
\STATE Prompt $S$ with one-shot $D$ ($U$+=1) 

--> sample with OneShotScore
\STATE OneShotScore - ZeroShotScore

--> sample with GoldenScore
\STATE Do score ranking and pick top K
\STATE \textbf{Return} Selected Subset
\end{algorithmic}
\end{algorithm}

\begin{algorithm}
\caption{SelecIT}
\label{alg:selectit}
\begin{algorithmic}[1]
\STATE \textbf{Init} $D$ = Candidate Dataset, $S$ = Backbone, $U$ = LLM Usage, $P$ = Prompt Templates
\STATE Use each prompt in $P$ to annotate each sample

--> Prompt-attached $D_p$
\STATE Use each model in $S$ to score each sample in $D_p$ ($U$+=3)

--> $D_p$ with score
\STATE Aggregate scores for each sample in $D$

--> each sample in $D$ has an aggregated score
\STATE Do score ranking and pick top K
\STATE \textbf{Return} Selected Subset
\end{algorithmic}
\end{algorithm}

\begin{algorithm}
\caption{IFD \& Superfiltering}
\label{alg:ifd+super}
\begin{algorithmic}[1]
\STATE \textbf{Init} $D$ = Candidate Dataset, $S$ = Backbone, $U$ = LLM Usage
\STATE Use $D' \in D$ to to warm up $S$ 

--> pre-experienced $S'$
\STATE Use $S'$ to generate IFD/Perplexity score on $D$ ($U$+=1) 

--> each sample with score
\STATE Do score ranking and pick top K
\STATE \textbf{Return} Selected Subset
\end{algorithmic}
\end{algorithm}

\begin{algorithm}
\caption{LESS}
\label{alg:less}
\begin{algorithmic}[1]
\STATE \textbf{Init} $D_c$ = Candidate Dataset, $D_t$ = Target Dataset, $S$ = Backbone, $U$ = LLM Usage
\STATE Use $D_c' \in D_c$ to LoRA warm up $S$ 

--> LoRA Model $S'$

\STATE Use $S'$ to get gradients of $D_c$ ($U$+=1) 

--> gradient store of $D_c$

\STATE Use $S'$ to get gradients of $D_t$ ($U$+=1) 

--> gradient store of $D_t$

\STATE Do gradient-similarity-based selection
\STATE \textbf{Return} Selected Subset
\end{algorithmic}
\end{algorithm}

\begin{algorithm}
\caption{DEITA}
\label{alg:deita}
\begin{algorithmic}[1]
\STATE \textbf{Init} $D$ = Candidate Dataset, $S$ = Backbone, $U$ = LLM Usage
\STATE Use evolved datasets to train two $S$s ($U$+=1)

--> complexity scorer model $S_c$ and quality scorer model $S_q$
\STATE Use $S_c$ to score $D$ ($U$+=1) 

--> instruction with complexity score
\STATE Use $S_q$ to score $D$ ($U$+=1) 

--> output with quality score
\STATE Multiply two scores and rank

--> ranked sample
\STATE Do score-first, diversity-aware selection ($U$+=1) 
\STATE \textbf{Return} Selected Subset
\end{algorithmic}
\end{algorithm}

\begin{algorithm}
\caption{InstructionGPT-4}
\label{alg:instructiongpt-4}
\begin{algorithmic}[1]
\STATE \textbf{Init} $D_c$ = Candidate Dataset, $D_t$ = Training Dataset, $S$ = Transformer model, $U$ = LLM Usage
\STATE Use vectorized $D_t$ to train a self-attention NN (U+=1)

--> trained $S$
\STATE Do vectorization on $D_c$ with indicators

--> vectorized $D_v$
\STATE Use $S$ to predict loss on $D_v$

--> sample with loss score

\STATE Do score ranking and pick top K
\STATE \textbf{Return} Selected Subset
\end{algorithmic}
\end{algorithm}

\begin{algorithm}
\caption{Instruction-Mining}
\label{alg:instruction-mining}
\begin{algorithmic}[1]
\STATE \textbf{Init} $D_c$ = Candidate Dataset, $D_t$ = Training Dataset, $S$ = Linear Selector, $U$ = LLM Usage
\STATE Use vectorized $D_t$ to train a linear selector

--> trained $S$
\STATE Do vectorization on $D_c$ with indicators

--> vectorized $D_v$
\STATE Use $S$ to predict loss on $D_v$

--> sample with loss score

\STATE Do score ranking and pick top K
\STATE \textbf{Return} Selected Subset
\end{algorithmic}
\end{algorithm}

\begin{algorithm}
\caption{SHED}
\label{alg:shed}
\begin{algorithmic}[1]
\STATE \textbf{Init} $D$ = Candidate Dataset, $S$ = Backbone, $U$ = LLM Usage
\STATE Do clustering on embedded $D$

--> Proxy dataset from each cluster $D_p$
\STATE Calculate Shapley Value on $D_p$

--> $D_p$ with score
\STATE Do optimization-aware sampling on $D$ according to scores

--> Selected Subset
\STATE \textbf{Return} Selected Subset
\end{algorithmic}
\end{algorithm}









\end{document}